\documentclass[sigconf,nonacm]{acmart}
\usepackage{multirow}
\usepackage{booktabs}
\usepackage{makecell}
\usepackage{tcolorbox}
\tcbuselibrary{breakable}
\usepackage{CJKutf8}
\usepackage{svg}

\AtBeginDocument{%
  }

\begin{document}

\title{Trust Your Memory: Verifiable Control of Smart Homes through Reinforcement Learning with Multi-dimensional Rewards}

\author{Kai-Yuan Guo}
\authornote{Both authors contributed equally to this research.}
\email{guoky16@midea.com}

\author{Jiang Wang}
\authornotemark[1]
\authornote{Corresponding author.}
\email{wangjiang50@midea.com}

\affiliation{%
  \institution{AI Research Center, Midea Group}
  \city{Shanghai}
  \country{China}
}

\author{Renjie Zhao}
\affiliation{%
  \institution{AI Research Center, Midea Group}
  \city{Shanghai}
  \country{China}
}
\email{zhaorj32@midea.com}

\author{Tianyi Wang}
\affiliation{%
  \institution{AI Research Center, Midea Group}
  \city{Shanghai}
  \country{China}
}
\email{wangty53@midea.com}

\author{Wandong Mao}
\affiliation{%
  \institution{AI Research Center, Midea Group}
  \city{Shanghai}
  \country{China}
}
\email{maowandong@midea.com}

\author{Yu Gao}
\authornotemark[2]
\affiliation{%
  \institution{AI Research Center, Midea Group}
  \city{Shanghai}
  \country{China}
}
\email{gaoyu11@midea.com}

\author{Mou Xiao Feng}
\authornotemark[2]
\affiliation{%
  \institution{AI Research Center, Midea Group}
  \city{Shanghai}
  \country{China}
}
\email{mouxf@midea.com}

\author{Yi Xu}
\authornotemark[2]
\affiliation{%
  \institution{AI Research Center, Midea Group}
  \city{Shanghai}
  \country{China}
}
\email{xuyi42@midea.com}

\renewcommand{\shortauthors}{Kai-Yuan Guo et al.}

\begin{abstract}
  Large Language Models (LLMs) have become a key foundation for enabling personalized smart home experiences. While existing studies have explored how smart home assistants understand user queries to control devices in real time, their ability to perform memory-driven device control remains challenging from both evaluation and methodological perspectives. In terms of evaluation, existing benchmarks either focus on immediate device control or general open-domain memory retrieval tasks, and therefore cannot effectively evaluate a model’s ability to perform memory-driven device control. Methodologically, while memory-driven device control can be approached using Reinforcement Learning, conventional RL methods generally rely on outcome-based supervision (i.e., whether the final task is achieved). This lack of intermediate feedback can lead to sub-optimal performance or local failures in fine-grained memory management tasks (adding, updating, deleting, and utilizing). To address these issues, we first release \textbf{MemHomeLife}, built from anonymized real-world long-term user interaction logs.
  To enable more fine-grained evaluation of different memory-related subtasks, we further construct \textbf{MemHome}, the first benchmark designed to systematically evaluate memory-driven device control in smart home scenarios. To meet the strict alignment requirements of MemHome, we further propose \textbf{Product Reward In Smart Memory (PRISM)}, a reinforcement learning alignment framework that combines multi-dimensional consistency rewards through multiplication. PRISM enforces strict supervision across multiple dimensions, where any violation in a single dimension (e.g., a format error) will invalidate the entire trajectory. Experiments demonstrate that PRISM significantly outperforms baseline methods (e.g., Mem0 and Memory-R1) on MemHome, achieving improvements of up to 15.43\% on our 8B-scale models. In a large-scale online A/B test spanning one month (involving millions of monthly user queries), our method improved the Task Completion Rate (TCR) of the smart home assistant by 1.34\%. The model and data will be open-sourced soon.
\end{abstract}

\ccsdesc[500]{Computing methodologies~Natural language processing}
\ccsdesc[500]{Computing methodologies~Reinforcement learning}

\keywords{LLMs, Memory, Agent, Reinforcement Learning, RLVR, Smart Home}

\maketitle

\begin{figure}[t]
  \centering
  \includegraphics[width=0.8\linewidth]{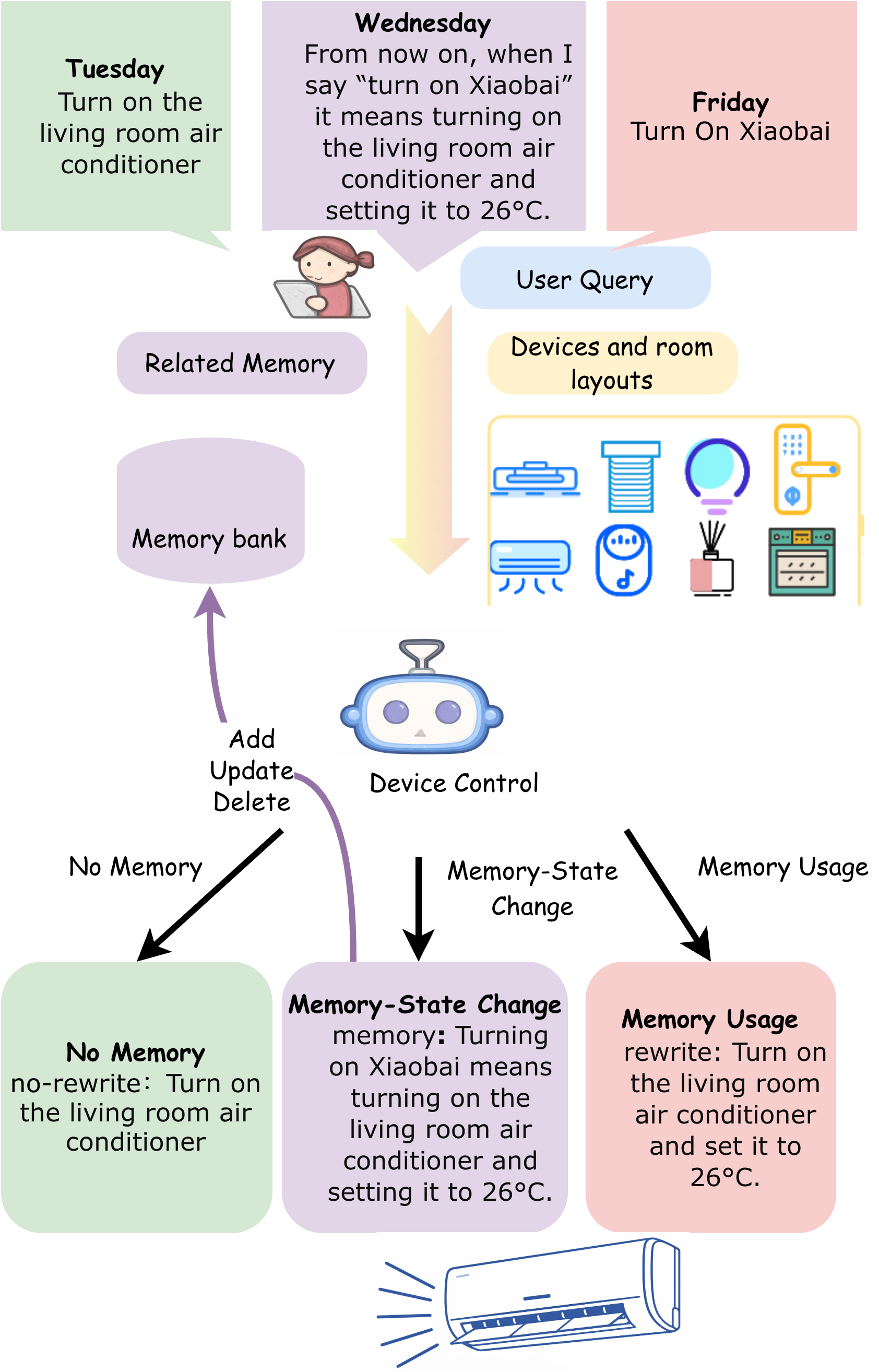}
  \caption{Overview of memory-driven device control framework in MemHome. Given a user query, rooms and layouts, the system retrieves relevant memories from the memory bank, and the device control model determines the appropriate control behavior. \textcolor[rgb]{0.71,0.77,0.71}{Green} indicates No-Memory control, where the instruction is executed directly. \textcolor[rgb]{0.75,0.71,0.77}{Purple} represents Memory State Change, where the model writes, updates, or deletes memory entries. \textcolor[rgb]{0.83,0.69,0.68}{Red} denotes Memory Usage, where the device control command is rewritten based on retrieved memory.}
  \label{fig:tast}
\end{figure}

\section{Introduction}

The introduction of Large Language Models (LLMs) has fundamentally transformed smart home assistants from rigid command executors into agents capable of understanding complex user intent. Although early studies \cite{king2024sasha,rivkin2024aiot,li2025homebench} have demonstrated the feasibility of using LLMs for device control, most existing approaches remain limited to a memory-free setting. This paradigm contradicts a core property of real-world home environments, which are highly personalized and persistent over time. In practice, an agent cannot simply map a user query to a command. Instead, it must incorporate users’ evolving habits and interaction history to generate device control commands that truly match user intent. As a result, a central challenge in smart home systems is enabling LLMs to perform memory-driven device control.

Smart home interactions involve both instantaneous user queries and long-term user memory. A system must determine, based on existing memory, which information should be added, updated, or removed, and must also correctly use memory in subsequent interactions to generate device control commands that align with user intent. Despite the importance of this capability, existing research lacks a systematic evaluation of a model’s ability to perform memory-driven device control. Current smart home benchmarks, such as HomeBench \cite{li2025homebench}, focus primarily on immediate device control in memory-free settings. In contrast, general memory benchmarks, such as LoCoMo \cite{maharana-etal-2024-evaluating} and LongMemEval \cite{wu2024longmemeval}, emphasize open-domain memory retrieval and mainly evaluate whether a model can correctly recall past information. We refer to this gap as the \textbf{Memory-to-Control Gap}: because existing benchmarks target different evaluation objectives, they fail to reveal systematic failures where a model recalls the correct information but generates an incorrect device control command.

To bridge this gap, we introduce \textbf{MemHome}, the first benchmark designed to systematically evaluate a model’s ability to perform memory-driven device control in smart home scenarios. MemHome is constructed from real user logs and enhanced with high logical density, and explicitly evaluates whether models can correctly handle memory-related behaviors such as adding, updating, deleting, and using memory when generating device control commands. This design shifts the evaluation focus from passive memory retrieval to active memory-driven control. In addition, we release \textbf{MemHomeLife}, a multi-turn interaction benchmark built from anonymized real-world data, which evaluates whether models can maintain memory correctness and complete device control tasks under long-term interactions and evolving memory.

Solving the tasks defined in our benchmarks poses unique challenges for existing alignment methods. The task requires an agent to simultaneously satisfy strict format constraints, memory logic, and device control rules, while producing outputs in natural language. We find that standard reinforcement learning with verifiable rewards (RLVR) \cite{shao2025deepseekmath,wang2025icpo,yan2025memory} is insufficient for this setting. Although the outputs have clear correctness criteria, they often depend on judgments at the natural language level and cannot be verified through direct execution. To address this limitation, some methods \cite{tang2025beyond} attempt to extend RLVR by using LLM-based feedback as reward signals. However, reward signals based solely on overall correctness still fail to capture critical fine-grained errors, such as incorrect room selection or confusion between device targets.

To address this issue, we propose \textbf{Product Reward In Smart Memory (PRISM)}, a multi-dimensional verifiable reward framework for natural language outputs. PRISM decomposes the correctness requirements of the task into multiple consistency dimensions, each of which is evaluated by an LLM-based verifier, and jointly uses their verification results as reward signals during training. If any single dimension, whether related to output format or memory logic, is violated, the reward for the entire generation trajectory is set to zero. Through this design, models receive strict feedback during training, enabling them to more effectively learn how to generate natural language outputs that satisfy both user intent and device control requirements. Our contributions are summarized as follows:
\begin{enumerate}
  \item We present MemHome and MemHomeLife as benchmarks designed to assess the accuracy of memory-based control command generation, effectively bridging the Memory-to-Control gap in current evaluation protocols.
  \item PRISM is proposed as a reinforcement learning reward framework that decomposes correctness requirements into multiple consistency dimensions and aligns them jointly during training.
  \item Extensive experiments show that PRISM significantly outperforms baseline models such as Mem0 \cite{chhikara2025mem0} and Memory-R1 \cite{yan2025memory}. Notably, our 4B-parameter model achieves performance comparable to a 72B general-purpose model, and large-scale online A/B tests demonstrate significant improvements in task completion rates in real-world settings. To facilitate further research, we will release all related resources (datasets and models) to the public.
\end{enumerate}

\section{Related Work}
\subsection{LLM-Based Smart Home Control}

LLMs have been introduced into smart home systems to improve the flexibility of natural language understanding and device control. For example, Sasha \cite{king2024sasha} focuses on parsing natural language instructions into executable device control actions. AIoT \cite{rivkin2024aiot} further treats LLMs as a central component and incorporates user profiles, historical interaction records, and tool-calling capabilities to support more complex tasks. Harmony \cite{yin2024harmony} and HomeBench \cite{li2025homebench} focus on multi-device scenarios, invalid instruction handling, and device command generation in complex environments, and construct corresponding benchmarks to evaluate models’ performance in immediate device control.

However, although some of these works (e.g., AIoT) may record user memory during interaction, such memory is used only as auxiliary context rather than being treated as core information that drives task execution. In contrast, our work takes memory-driven device control as the central evaluation target and focuses on assessing whether models can correctly use memory to generate device control commands that align with user intent.

\subsection{Memory-Related Methods and Benchmarks}

Recent work has shown that LLM-based agents need to handle memory during interaction, including deciding what to record, update, delete, and use. Early systems such as MemoryBank \cite{zhong2024memorybank}, MemGPT \cite{packer2023memgpt}, and Mem0 \cite{chhikara2025mem0} improve access to past information through external memory structures and retrieval mechanisms. Building on this line of work, Mem-$\alpha$ \cite{wang2025mem} and Memory-R1 \cite{yan2025memory} further treat memory handling as a learnable process and apply reinforcement learning, where rewards are primarily defined by exact-match correctness on final question answering tasks. However, these methods mainly optimize for information recall.

As LLMs become capable of processing long contexts, evaluating memory-related behavior has emerged as a separate research direction. Early benchmarks such as LoCoMo \cite{maharana-etal-2024-evaluating} and LongMemEval \cite{wu2024longmemeval} focus on whether models can correctly recall historical information through question answering. Later benchmarks begin to examine memory use during interaction. Mem-PAL \cite{huang2025mem} evaluates personalized response generation based on dialogue and behavior history. MemBench \cite{tan-etal-2025-membench} evaluates whether models can correctly remember and answer questions after long user–assistant interactions. More recently, MemoryBench \cite{ai2025memorybench} combines memory with task execution by simulating user feedback and evaluating whether models can adjust their behavior over time.

Despite these efforts, existing benchmarks treat memory either as the evaluation target itself for question answering or as auxiliary information for performance improvement. Memory is not treated as information that directly drives task execution, where models must rely on memory to process user queries and complete tasks. Moreover, these benchmarks do not clearly distinguish different types of memory-related behaviors, such as whether information should be recorded or how memory should be used.

In contrast, we focus on smart home assistants, where models must generate device control commands based on memory. We introduce MemHome to systematically evaluate memory-driven device control in smart home scenarios. MemHome evaluates whether models can decide when memory should be recorded, and whether they can correctly use memory to generate device control commands aligned with users’ implicit intent. By grouping interactions into memory state changes, memory usage, and cases where memory is not triggered, MemHome enables more precise evaluation and provides clearer signals for reinforcement learning.

\subsection{Reward Design in Reinforcement Learning}

In reinforcement learning, LLM alignment depends on reward design. A common approach is reinforcement learning with verifiable rewards (RLVR), which has been effective in tasks such as mathematical reasoning and code generation \cite{shao2025deepseekmath,team2025kimi}, where correctness can be verified through executable answers. To apply RLVR to natural language tasks, recent work uses intrinsic probability signals \cite{yu2025rlpr,wang2025icpo,tang2025beyond} or LLM-based verification \cite{su2025crossing,10.5555/3692070.3693141,peng-etal-2025-verif} as rewards. However, relaxing verification also weakens correctness checking, making it difficult to identify fine-grained semantic or key information errors in tasks with strict correctness requirements.

In contrast, PRISM emphasizes strict consistency. In smart home scenarios, each user query has a clear correct outcome, but verification is performed over free-form text. PRISM decomposes correctness into multiple consistency dimensions and applies a multiplicative reward, making it better suited for memory-driven device control.

\section{Task Definition}

We formulate our task as follows. At interaction step $t$, the system generates an output $y_t$ based on the user query $q_t$, dialogue history $d_t$, room layouts and devices $c_t$, and the memory state $M_t$. We use $\pi_\theta$ to denote the device control policy parameterized by $\theta$.
\begin{align}
  \pi_\theta : (q_t, d_t, c_t, M_t) \rightarrow y_t, M_{t+1}
\end{align}
If the output $y_t$ contains information indicating memory writing, updating, or deletion, the memory state $M_{t}$ is updated to $M_{t+1}$; otherwise, $M_{t+1} = M_t$, and the output $y_t$ is passed to subsequent processing.

\section{System Framework}
In MemHome, each user query is processed by a pipeline consisting of a device control model and a memory bank. Figure \ref{fig:tast} illustrates this pipeline. The device control model $\pi_\theta$ generates output with an explicit prefix to indicate the memory-related behavior of the current interaction. Specifically, different behaviors correspond to different predefined output prefixes: (i) for memory writing or updating, the model outputs "memory: <content>", and for memory deletion, it outputs "memory: delete <content>"; (ii) for memory-based rewriting, it outputs "rewrite: <rewritten device control command>";
(iii) when memory is not involved, the model directly outputs
"no-rewrite".
The memory bank executes memory updates and retrieval. When the device control model outputs a memory modification command with prefix "memory", the memory bank performs the corresponding write, update, or deletion operation.

\subsection{Execution Flow}

Upon receiving a user query, the system first retrieves relevant memories from the memory bank. The device control model then takes the user query, room layouts and devices, and retrieved memories as input, and determines whether the user query should modify the memory bank, whether existing memory should be used to rewrite the query, or whether the query should be handled without involving memory. The output is routed to the memory bank to execute the corresponding memory operation only when the device control model determines that the memory state should be changed, as indicated by the output prefix “memory”. Otherwise, the system proceeds to subsequent modules for executing the corresponding device control commands. This design is motivated by the observation that most user queries do not modify memory. Invoking the memory bank for every query would incur unnecessary overhead and latency; therefore, it is activated only when the model explicitly determines that a memory update is required.

\section{MemHome Dataset}
\subsection{Data construction}

In real-world scenarios, users often express, revise, or revoke memory across days and sessions, while the system is required to interpret user intent and generate device control commands under an evolving memory state. Each raw interaction sequence naturally contains three types of information: (1) devices and room layouts; (2) the evolving memory state; and (3) user–assistant dialogues.

We first manually curate and annotate anonymized long-term interaction logs. For each interaction turn, we explicitly annotate memory state changes, identify which memory entries the user query depends on, and label the corresponding task type and ground-truth output. This process results in \textbf{MemHomeLife}, which is designed to evaluate model performance under long-term interactions with evolving memory.

However, in such real long-term interactions, multiple memory-related behaviors, such as memory writing, updating, deletion, and usage, are often intertwined across dialogue turns. While this structure faithfully reflects practical scenarios, it makes it difficult to isolate and analyze a model’s capability with respect to a specific behavior, such as whether existing memory is correctly used.

\begin{figure}[t]
  \centering
  \includegraphics[width=1\linewidth]{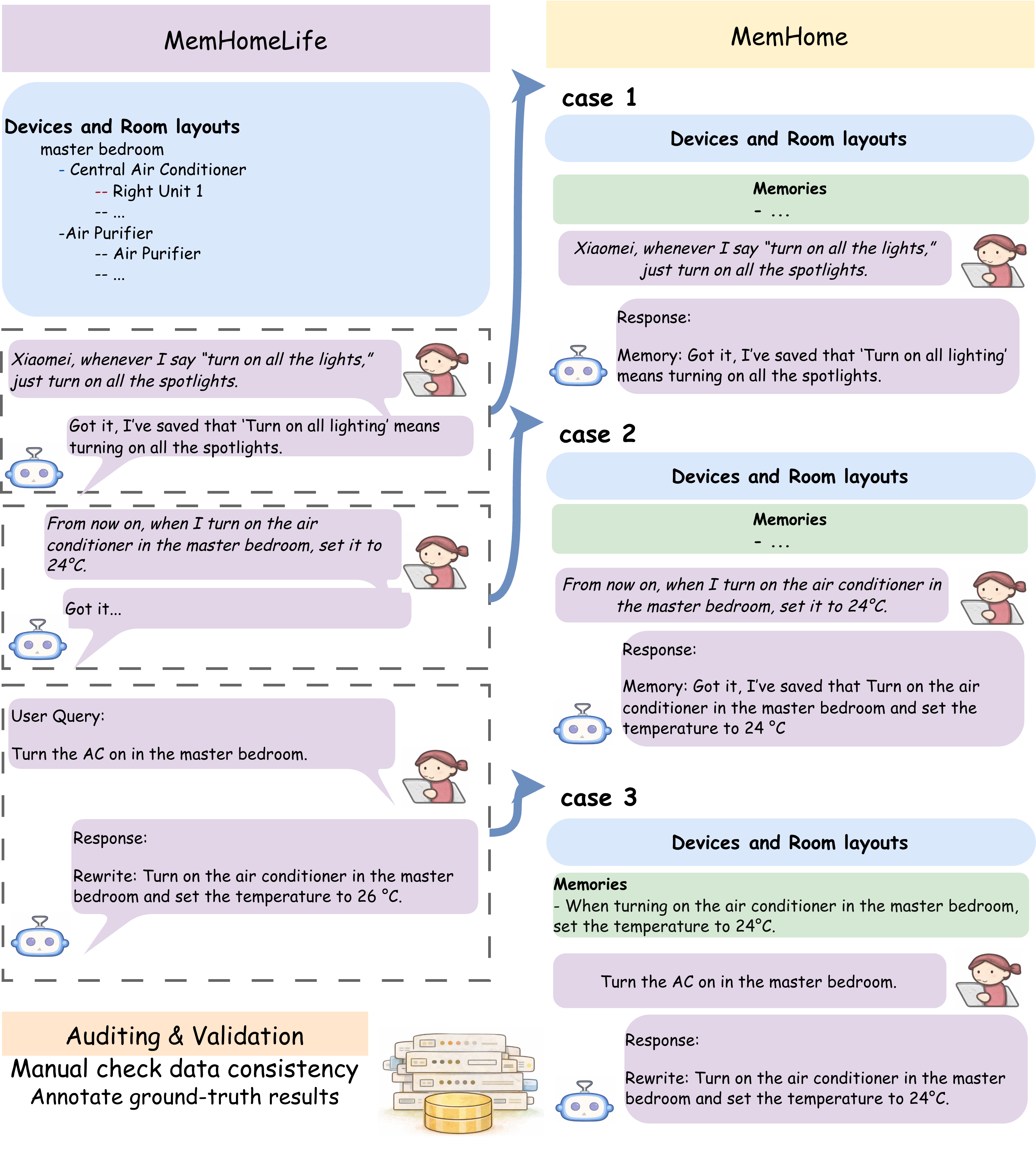}
  \caption{MemHome: From Real Long-Term Dialogues to Memory-Centric Evaluation Cases. MemHomeLife preserves multi-turn interactions with evolving memory states, while MemHome is constructed by extracting cases with explicitly annotated relevant memories under a fixed environment.}
  \label{fig:dataset}
\end{figure}

To address this issue, we further construct \textbf{MemHome}, a memory-driven device control benchmark for fine-grained evaluation. Specifically, we segment the data by session, fixing the devices and room layouts at the beginning of each session and retaining the in-session dialogue history as short-term context. For each session, we select the final user query as the evaluation target and assemble a set of candidate memory entries, including both the target and non-target memory entries, according to manual annotations. Each MemHome sample thus corresponds to an independent task of generating a device control command given a user query and a set of candidate memories. To further increase evaluation difficulty, we additionally curate challenging test cases from long-term logs where users explicitly expressed dissatisfaction with system responses, such as scenarios with multiple competing memories and colloquial user expressions. Through the complementary design of MemHomeLife and MemHome, we enable both holistic evaluation under long-term interactions and fine-grained analysis of different memory-related behaviors, while preserving the original data distribution and memory evolution logic. The construction process of MemHome is illustrated in Figure \ref{fig:dataset}.

\textbf{Human Annotation and Quality Control} To ensure the reliability of evaluation results, all labels in MemHome and MemHomeLife are manually annotated and cross-validated. In addition, the devices and room layouts associated with each dialogue is manually checked for consistency, ensuring that there are no logical conflicts among the dialogue flow, devices, and memory states.

\subsection{Dataset Comparison and Analysis}

\begin{table*}[t]
  \centering
  \small
  \setlength{\tabcolsep}{6pt}
  \renewcommand{\arraystretch}{1.15}
  \begin{tabular}{lcccc}
    \toprule
    \textbf{Property} &
    \textbf{HomeBench} &
    \textbf{LoCoMo} &
    \textbf{MemoryBench} &
    \textbf{MemHome (Ours)} \\
    \midrule
    Application Domain
    & Smart Home
    & Open-domain
    & General Agent
    & Smart Home \\

    Memory-related Task
    & $\times$ & $\checkmark$ & $\checkmark$ & $\checkmark$ \\

    Explicit Memory Subtasks
    & $\times$ & $\times$ & $\times$ & $\checkmark$ \\

    Instruction Tuning
    & $\checkmark$ & $\times$ & $\times$ & $\checkmark$ \\

    Language
    & En & En & En & ZH \\
    \bottomrule
  \end{tabular}
  \caption{Comparison of MemHome with existing related benchmarks.}
  \label{tab:dataset_comparison}
\end{table*}

As shown in Table~\ref{tab:dataset_comparison}, while prior benchmarks cover smart home interaction or memory-related tasks in isolation, they do not explicitly evaluate memory as key information for device control. MemHome and MemHomeLife differ in that they explicitly define memory-related subtasks and require models to generate memory-driven device control commands that are strictly aligned with user intent.

MemHome contains a total of $1945$ samples. We use $80\%$ of the data for training and reserve the remaining $20\%$ (389 samples) as a held-out evaluation set for all models. Table \ref{tab:memhome_stats} reports statistics of MemHome evaluation set across different categories in terms of environment complexity and context scale.

\textbf{Memory Use}. This category requires models to correctly retrieve and use existing memory to interpret user intent and rewrite the user query accordingly. The evaluation set includes $220$ samples in this category, capturing the most common usage scenario in daily smart home interactions where memory is actively used for understanding and execution.

\textbf{No-Memory}. This category includes both No-Memory and Do-Not-Memorize cases, with a total of $53$ evaluation samples. In these scenarios, the current query should neither trigger any memory modification nor rely on existing memory for rewriting. Models are expected to directly interpret and execute the query, and this category evaluates a model’s ability to avoid both inappropriate memory usage and unnecessary memory updates.

\textbf{Memory State Change}. This category includes memory addition and memory deletion, with $116$ evaluation samples. These cases characterize interactions that require modifying the memory state. Models must correctly determine when user intent should be written into memory or when existing memory should be removed if it is no longer needed.

\begin{table}[h]
  \centering
  \footnotesize
  \setlength{\tabcolsep}{4pt}
  \renewcommand{\arraystretch}{1.15}
  \begin{tabular}{lcccc}
    \toprule
    & \textbf{No-Mem} & \textbf{Mem-Use} & \textbf{Mem-Change} & \textbf{All} \\
    \midrule
    \#Samples
    & 53 & 220 & 116 & 389 \\

    Rooms
    & 15.06/16
    & 14.77/16
    & 14.87/16
    & 14.84/16 \\

    Devices
    & 118.91/124
    & 116.52/129
    & 117.07/129
    & 117.01/129 \\

    Hist. Turns
    & 1.02/5
    & 0.76/3
    & 0.85/4
    & 0.82/5 \\

    Memory
    & 2.06/3
    & 1.10/2
    & 1.78/4
    & 1.43/4 \\
    \bottomrule
  \end{tabular}
  \caption{Statistics of the MemHome evaluation set. Values are reported as avg/max.}
  \label{tab:memhome_stats}
\end{table}

We further show the statistics of MemHomeLife. On average, each dialogue contains $14.3$ interaction turns, with a maximum of $33$ turns. The dialogue span approximately $11$ days on average and up to $26$ days at most, demonstrating clear cross-day long-term interaction characteristics. Each dialogue involves an average of $3.8$ memory entries, with a maximum of seven, and spans an average of $7.2$ distinct sessions, reaching up to $13$ sessions in the longest cases.

\section{Evaluation Metrics}

Under both the MemHome and MemHomeLife settings, we evaluate models based on the final output generated at the last interaction turn of each sample. For Memory-Use and Memory-State-Change samples, we evaluate whether models generate device control commands aligned with user intent.

\textbf{Text Similarity Metrics}. We use token-level F1 and BLEU-1 (B1) to measure the similarity between the generated device control command and the ground-truth. For Chinese instructions, tokens correspond to individual Chinese characters. Detailed computation procedures are provided in the appendix \ref{similarity}. However, in memory-driven device control, correctness is highly sensitive to key information, and even minor errors in device targets or control parameters can invalidate the entire command, which cannot be reliably captured by token-level F1 or B1 based similarity metrics.

\textbf{LLM-Based Consistency Judgment (LLM-as-Judge).}
We mainly use an LLM-based judgment to evaluate whether model outputs are semantically consistent with the ground-truth. Specifically, two independent judge models provide judgments (Yes or No) on whether the model output is semantically consistent with the ground-truth; when they disagree, a third model is used to produce the final judgment. The detailed judgment prompts are provided in appendix \ref{prompts}.

Content-level evaluation is applied only to Memory Use and Memory State Change samples, where models are required to generate device control commands. For No-Memory samples, content-level evaluation is not performed.

\section{Method}

In MemHome, each user query has a well-defined correct outcome. While reinforcement learning with verifiable rewards (RLVR) is effective for tasks with exact answers, such as code generation, in our task the ground truth is expressed in natural language, where existing extensions of RLVR typically rely on coarse-grained verification and lack the ability to identify fine-grained errors in device selection, parameter specification, and other task-critical details.

\begin{figure}[h]
  \centering
  \includegraphics[width=0.8\linewidth]{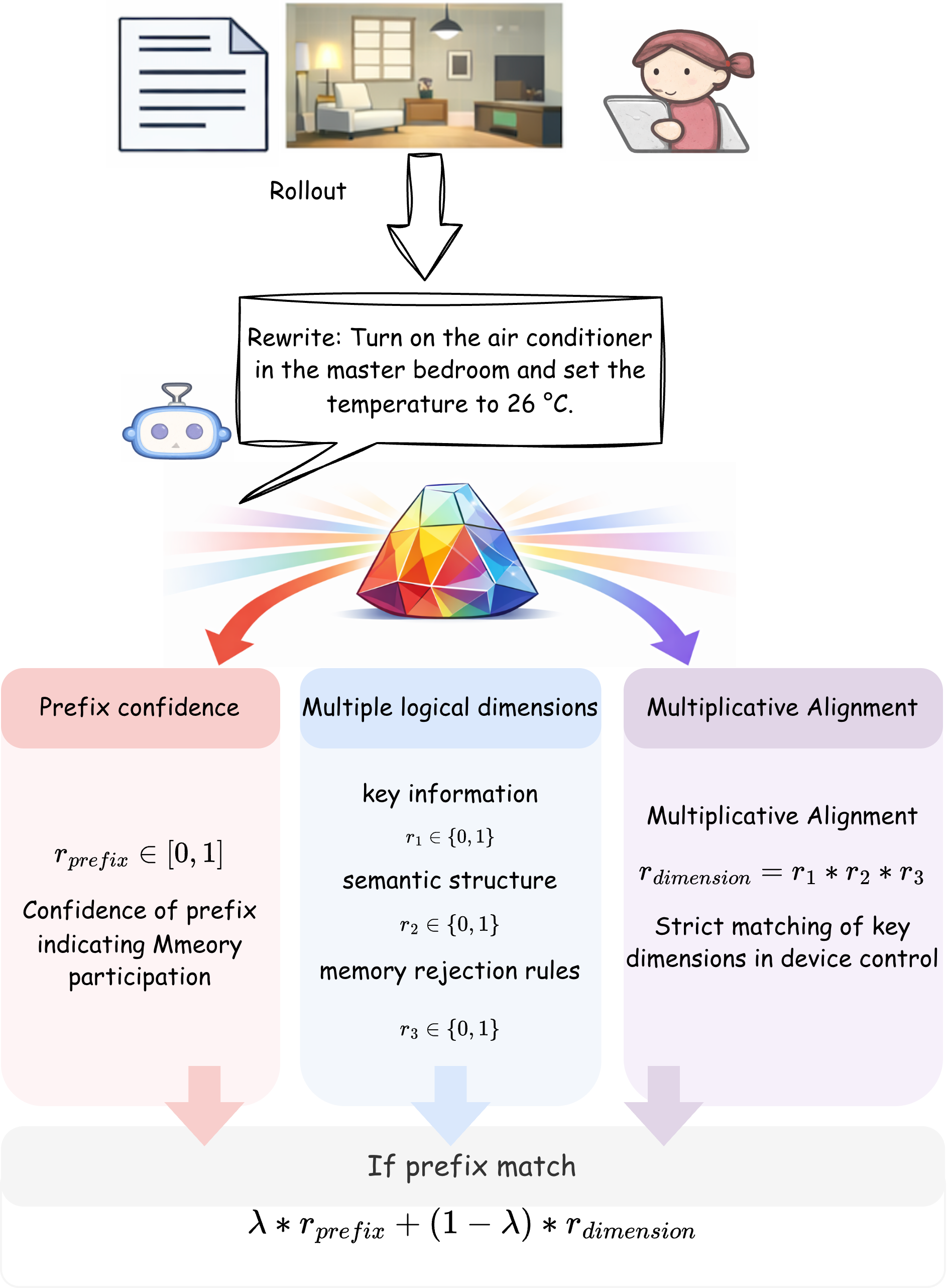}
  \Description{Overview of the PRISM reward design, showing prefix confidence, multiple consistency checks, and a multiplicative veto-based reward aggregation.}
  \caption{Overview of the Product Reward In Smart Memory.}
  \label{fig:reward}
\end{figure}

To address this limitation, we propose PRISM, a reinforcement learning framework that enforces strict multi-dimensional alignment via a veto-based multiplicative reward. An overview of PRISM is illustrated in Figure \ref{fig:reward}.

\subsection{Multi-Dimensional Consistency with Veto-Based Reward}

We decompose output correctness into three consistency dimensions that capture the core requirements of memory-driven device control:
(1) \textbf{Key information consistency.} It verifies whether the generated output matches the ground truth in essential control semantics, such as the target device, operation type, parameters, and execution scope (e.g., distinguishing “turn on the air conditioner” from “turn on the living-room air conditioner” when room specification is required).
(2) \textbf{Semantic structure and intent consistency.} It assesses whether the output preserves the user’s intended control logic, avoiding unintended structural shifts such as converting a compositional action into a conditional trigger or vice versa (e.g., “when turning on the washing machine, set the water temperature to 40°C” vs. “turn on the washing machine and set the water temperature to 40°C”).
(3) \textbf{Memory rejection consistency.} It focuses on incorrect memory writing or usage where memory should not be written or applied, ensuring that the model correctly rejects memory updates for non-device-related or sensitive information (e.g., requests to memorize personal identifiers).
The detailed judgment prompts are provided in the appendix \ref{prompts}. For each dimension $k$, an LLM produces a binary judgment $r_k \in \{0,1\}$ indicating whether the output matches the ground truth on that dimension. PRISM aggregates these judgments using a veto-based multiplicative reward structure:
\begin{align}
  r_{\text{dimension}} = \prod_{k=1}^{K} r_k .
\end{align}
A positive reward is obtained only when the output is consistent across all dimensions; any single inconsistency vetoes the entire reward, preventing partially correct outputs from being rewarded. This veto-based design also mitigates reward hacking, where models may exploit the reward by generating outputs that appear correct but contain critical errors in key information. By requiring all consistency dimensions to be satisfied simultaneously, PRISM prevents partially correct generations (e.g., correct format but incorrect device or parameters) from receiving positive rewards.

\subsection{Prefix Confidence Reward}
To encourage more reliable task-type prediction during training, we further introduce a prefix confidence reward as an auxiliary signal, which captures the model’s confidence in generating the correct prefix (e.g., memory, rewrite, no-rewrite). During forward, we extract the token-level log-probabilities corresponding to the ground-truth prefix and aggregate them to estimate the likelihood of generating the correct prefix:
\begin{align}
  p_{\text{pfx}} = \exp\!\left( \sum_{t \in \mathcal{P}} \log p_\theta(t \mid \text{context}) \right),
\end{align}
where $\mathcal{P}$ denotes the set of tokens in the ground-truth prefix. This value reflects the model’s overall tendency to generate the correct prefix. We apply the following transformation to map it into the $(0,1)$ interval:
\begin{align}
  \text{logit}_{\text{pfx}} &= \log \frac{p_{\text{pfx}}}{1 - p_{\text{pfx}}}, \\
  r_{\text{prefix}} &= \frac{\tanh(\text{logit}_{\text{pfx}}) + 1}{2}.
\end{align}
This makes the prefix confidence suitable for combination with other reward terms.

\subsection{Reward Composition}
The final reward is computed only when the predicted prefix matches the ground-truth. Otherwise, the reward is set to $0$. This design enforces that the model first learns to correctly identify the task type. For samples with a correct prefix, the final reward is defined as
\begin{align}
  r =
  \begin{cases}
    \lambda \, r_{\text{prefix}} + (1 - \lambda) \, r_{\text{dimension}}, & \text{if prefix matches}, \\
    0, & \text{otherwise}.
  \end{cases}
\end{align}
where $r_{\text{dimension}}$ enforces strict correctness across all dimensions, and $r_{\text{prefix}}$ serves as an auxiliary signal.

\section{Experiment}

\begin{table*}[t]
  \centering
  \small
  \setlength{\tabcolsep}{5pt}
  \renewcommand{\arraystretch}{1.25}
  \begin{tabular}{cc|ccc|ccc|ccc|ccc}
    \hline
    \multirow{2}{*}{\textbf{Base Model}}& \multirow{2}{*}{\textbf{Method}}
    & \multicolumn{3}{c|}{\textbf{No-Memory}}
    & \multicolumn{3}{c|}{\textbf{Memory Use}}
    & \multicolumn{3}{c|}{\textbf{Memory State Change}}
    & \multicolumn{3}{c}{\textbf{Overall}} \\
    &
    & F1$\uparrow$ & B1$\uparrow$ & Acc$\uparrow$
    & F1$\uparrow$ & B1$\uparrow$ & Acc$\uparrow$
    & F1$\uparrow$ & B1$\uparrow$ & Acc$\uparrow$
    & F1$\uparrow$ & B1$\uparrow$ & Acc$\uparrow$ \\
    \hline

    \multirow{4}{*}{\textbf{Qwen3-4B}}
    & Mem0
    & 41.82 & 38.86 & 37.74
    & 50.70 & 47.20 & 34.55
    & 77.94 & 73.93 & 68.97
    & 57.61 & 54.03 & 45.25 \\

    & Memory-SFT& 39.82 & 36.77 & 35.85
    & 58.55 & 57.97 & 57.73
    & 61.84 & 58.27 & 50.86
    & 56.98 & 55.17 & 52.70 \\

    & Memory-R1
    & 41.64 & 38.66 & 41.64
    & 49.96 & 47.71 & 42.73
    & 80.96 & 76.94 & 72.41
    & 58.07 & 55.19 & 51.43 \\

    & PRISM
    & \textbf{56.88} & \textbf{54.25} & \textbf{52.83}
    & \textbf{65.10} & \textbf{62.85} & \textbf{58.64}
    & \textbf{81.30} & \textbf{76.94} & \textbf{72.41}
    & \textbf{68.81} & \textbf{65.88} & \textbf{61.95} \\

    \hline

    \multirow{4}{*}{\textbf{Qwen3-8B}}
    & Mem0
    & 43.09 & 41.39 & 52.83
    & 57.61 & 56.22 & 55.00
    & \textbf{72.63} & \textbf{69.29} & 68.10
    & 60.11 & 58.10 & 58.61 \\

    & Memory-SFT& 47.00 & 45.44 & 60.38
    & 72.26 & 70.78 & 65.00
    & 68.92 & 57.22 & 74.14
    & 67.82 & 63.28 & 67.10 \\

    & Memory-R1
    & 36.48 & 34.40 & 41.51
    & 69.74 & 68.52 & 69.74
    & 71.84 & 59.72 & 71.84
    & 65.83 & 61.25 & 66.52 \\

    & PRISM
    & \textbf{58.48} & \textbf{57.37} & \textbf{67.92}
    & \textbf{73.61} & \textbf{71.81} & \textbf{74.55}
    & 70.32 & 60.75 & \textbf{75.86}
    & \textbf{70.57} & \textbf{66.54} & \textbf{74.04} \\

    \hline
  \end{tabular}
  \caption{Main results of different training strategies on MemHome.
    We report F1, BLEU-1 (B1), and LLM-as-Judge accuracy (Acc) for each subtask category and overall.
  All metrics are reported as percentages (\%) and are higher-is-better. Best results are highlighted in bold.}
  \label{tab:memhome_4b_8b}
\end{table*}

\begin{table*}[t]
  \centering
  \small
  \setlength{\tabcolsep}{5pt}
  \renewcommand{\arraystretch}{1.25}
  \begin{tabular}{l|ccc|ccc|ccc|ccc}
    \hline
    \multirow{2}{*}{\textbf{Model}}
    & \multicolumn{3}{c|}{\textbf{No-Memory}}
    & \multicolumn{3}{c|}{\textbf{Memory Use}}
    & \multicolumn{3}{c|}{\textbf{Memory State Change}}
    & \multicolumn{3}{c}{\textbf{Overall}} \\
    & F1$\uparrow$ & B1$\uparrow$ & Acc$\uparrow$
    & F1$\uparrow$ & B1$\uparrow$ & Acc$\uparrow$
    & F1$\uparrow$ & B1$\uparrow$ & Acc$\uparrow$
    & F1$\uparrow$ & B1$\uparrow$ & Acc$\uparrow$ \\
    \hline

    GPT-5 & \textbf{61.82} & \textbf{60.99} & \textbf{73.58}
    & 69.42 & 68.37 & 70.00
    & 78.85 & 75.37 & 79.31
    & 71.20 & 69.45 & 73.26 \\

    Gemini-2.5 Pro & 58.11 & 57.25 & \textbf{73.58}
    & \textbf{78.35} & \textbf{76.91} & \textbf{77.73}
    & \textbf{85.02} & \textbf{81.16} & \textbf{82.76}
    & \textbf{77.58} & \textbf{75.50} & \textbf{78.66} \\

    Deepseek-V3.2 & 58.38 & 57.51 & 71.70
    & 63.24 & 61.69 & 61.36
    & 80.21 & 76.34 & 75.00
    & 67.64 & 65.49 & 66.84 \\

    Qwen3-235B & 50.50 & 49.67 & 66.04
    & 70.57 & 68.81 & 65.45
    & 80.11 & 76.59 & 74.14
    & 70.68 & 68.52 & 68.12 \\

    Qwen2.5-72B & 46.61 & 45.62 & 56.60
    & 64.12 & 62.17 & 60.45
    & 84.86 & 81.22 & 76.72
    & 67.92 & 65.60 & 64.78 \\

    \hline
  \end{tabular}
  \caption{Main results of different device control models on MemHome.
  We report F1, BLEU-1 (B1), and LLM-as-Judge accuracy (Acc) for each subtask category as well as overall. All metrics are reported as percentages (\%) and are higher-is-better. Best results are highlighted in bold.}
  \label{tab:memhome_answer_main}
\end{table*}

\subsection{Experimental Setup}
\textbf{Models and Baselines}. The first set of experiments focuses on small-scale open-source models (Qwen3-4B and Qwen3-8B) to assess the effectiveness of different fine-tuning strategies. Specifically, we consider four settings:
(1) Mem0 \cite{chhikara2025mem0}, a modular memory system that explicitly supports memory read, write, and update operations.
(2) Memory-SFT, a supervised fine-tuning baseline trained on the same MemHome data.
(3) Memory-R1, a model aligned via reinforcement learning using exact-matching reward designs from \cite{yan2025memory}.
(4) Ours, a model aligned via reinforcement learning using the proposed PRISM reward.

The second set of experiments aims to identify the empirical performance ceiling of current models on this task. We evaluate multiple mainstream models, all used directly for inference under a few-shot setting without any fine-tuning.

\textbf{Input and Inference Protocol}. All models receive the same inputs, including devices and room layouts, the current memory state, the dialogue history, and the user query. All models generate a device control command in response to the user query.

\textbf{Training Details and Evaluation.} The reinforcement learning stage is implemented using the GRPO framework \cite{shao2024deepseekmath}. Reward signals are applied only to the final token of each output sequence. Other training hyperparameters and computational settings are provided in the appendix \ref{parameters}. It is worth noting that the LLM-based consistency judgments used for reward modeling during training are independent of the LLM-as-Judge employed during evaluation.

\subsection{Main Results}

We first evaluate the effectiveness of PRISM by comparing different training strategies on Qwen3-4B and Qwen3-8B models, as shown in Table~\ref{tab:memhome_4b_8b}. On both model sizes, PRISM consistently outperforms strong baselines, including Mem0 and Memory-R1. In particular, our 8B-scale model achieves an overall accuracy improvement of up to $15.43\%$ over Mem0. These gains are consistently observed across all three memory-related sub-tasks, with improvements of $+15.09\%$ on No-Memory, $+19.55\%$ on Memory Use, and $+7.76\%$ on Memory State Change, while Memory-R1 yields smaller and less consistent improvements. These results indicate that, on the MemHome task, the exact-match reward of final answer used in Memory-R1 is less effective at improving task performance than the proposed multi-dimensional consistency reward in PRISM. In the Memory State Change sub-task on the 8B-scale model, PRISM does not achieve the highest scores on F1 or BLEU-1. We attribute this to the limitations of surface-level text similarity metrics in capturing correctness for memory-driven device control.

Performance on MemHome is further examined across several large language models, including GPT-5, Gemini-2.5 Pro, Qwen3-235B-A22B \cite{yang2025qwen3}, DeepSeek-v3 \cite{liu2024deepseek}, and Qwen2.5-72B \cite{qwen2.5}, with results shown in Table \ref{tab:memhome_answer_main}. Overall, these models achieve accuracies ranging from $64.78\%$ to $78.66\%$. Notably, our 8B-scale model achieves $94\%$ of the accuracy of the best-performing large-scale model (Gemini2.5 Pro), despite its substantially smaller model size. Nevertheless, performance across these models varies substantially across memory-related sub-tasks, and no single model consistently achieves the best results across all categories, highlighting the challenges posed by memory-driven device control beyond standard instruction understanding or factual recall.

Additional evaluation on MemHomeLife assesses performance under long-term interaction settings. Results show that our PRISM-aligned 8B-scale model achieves an overall LLM-as-Judge accuracy of $78.04\%$, as reported in Tables \ref{tab:memhome_e2e_llm} and \ref{tab:memhome_e2e_small}. The observed trends are consistent with those on MemHome, and are not discussed further here.

\begin{table}[h]
  \centering
  \small
  \setlength{\tabcolsep}{6pt}
  \renewcommand{\arraystretch}{1.25}
  \begin{tabular}{l|ccc}
    \hline
    \textbf{Model}
    & \textbf{F1$\uparrow$}
    & \textbf{BLEU-1$\uparrow$}
    & \textbf{Acc$\uparrow$} \\
    \hline

    GPT-5
    & 72.04 & 66.90 & 64.17 \\

    Gemini-2.5 Pro
    & \textbf{72.36} & \textbf{68.34} & \textbf{68.33} \\

    Deepseek-V3.2
    & 67.46 & 61.97 & 55.83 \\

    Qwen3-235B
    & 62.70 & 56.84 & 53.33 \\

    \hline
  \end{tabular}
  \caption{Main results of different device control models on MemHomeLife.}
  \label{tab:memhome_e2e_llm}
\end{table}

\begin{table}[h]
  \centering
  \small
  \setlength{\tabcolsep}{6pt}
  \renewcommand{\arraystretch}{1.25}
  \begin{tabular}{cc|ccc}
    \hline
    \textbf{Base model}
    & \textbf{Method}
    & \textbf{F1$\uparrow$}
    & \textbf{BLEU-1$\uparrow$}
    & \textbf{Acc$\uparrow$} \\
    \hline

    \multirow{4}{*}{\textbf{Qwen3-4B}}& Mem0
    & 38.22 & 37.36 & 35.00 \\

    & Memory-SFT
    & 44.02 & 42.36 & 40.00 \\

    & Memory-R1
    & 43.55 & 40.17 & 33.33 \\

    & Prism
    & \textbf{53.22} & \textbf{50.21} & \textbf{48.33} \\

    \hline

    \multirow{4}{*}{\textbf{Qwen3-8B}}& Mem0
    & 49.15 & 47.44 & 45.83 \\

    & Memory-SFT
    & 53.07 & 48.32 & 49.17 \\

    & Memory-R1
    & 57.06& 51.49& 50.83 \\

    & Prism
    & \textbf{60.91}& \textbf{54.93}& \textbf{53.33} \\

    \hline
  \end{tabular}
  \caption{Main results of different training strategies on MemHomeLife.}
  \label{tab:memhome_e2e_small}
\end{table}

\subsection{Ablation Study}
\begin{figure}[h]
  \centering
  \includegraphics[width=1\linewidth]{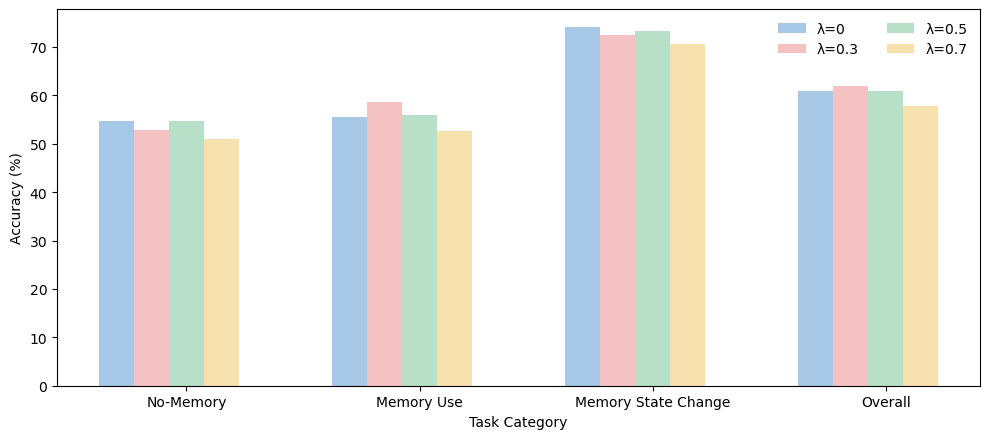}
  \caption{Ablation study on reward design with Qwen3-4B. We compare different prefix reward weights $\lambda$.}
  \label{fig:ablation}
\end{figure}

To analyze the contribution of each component in the reward design, we conduct ablation studies on the PRISM(4B) model by varying the reward configurations on MemHome, while keeping the training data and optimization steps fixed. The results are summarized in Figure~\ref{fig:ablation}. In general, introducing the prefix confidence reward consistently improves performance across memory-related tasks. A moderate weight ($\lambda = 0.3$) yields the best performance, improving the overall score by approximately $1\%$ compared to removing the prefix reward ($\lambda = 0$) and by about $3\%$ relative to an extreme weighting ($\lambda = 0.7$). Although $\lambda = 0.3$ does not achieve the highest score on every individual sub-task, it provides the most balanced performance across categories, supporting our design choice of treating the prefix confidence as an auxiliary rather than dominant training signal.

We also observe that replacing the veto-based multiplicative reward with an additive multi-dimensional formulation ($r_{\text{dimension}} = \sum_k r_k$) leads to a $4.62\%$ drop in overall accuracy. Detailed results are provided in the appendix \ref{ablation}.

\subsection{Online Deployment Results}
We deployed PRISM into a real-world smart home assistant system to handle device control tasks in practical scenarios. After demonstrating stable and consistent improvements over strong baselines in offline evaluations on MemHome and MemHomeLife, we conducted a large-scale online A/B test spanning one month, involving millions of real user queries. The results show that our method improved the overall Task Completion Rate (TCR) of the smart home assistant by $1.34\%$. In addition, we observed a noticeable reduction in memory-related execution errors, indicating more reliable memory-driven device control behavior in real-world usage.

\section{Conclusion}
This paper introduces MemHome and MemHomeLife, benchmarks for evaluating memory-driven device control in smart home. These benchmarks address a critical gap in existing evaluations, where memory is rarely assessed as a core information of device control tasks. We further propose PRISM, a reinforcement learning alignment framework that enforces strict correctness by decomposing correctness requirements into multiple consistency dimensions and jointly aligning them during training. Offline experiments on MemHome and MemHomeLife show that PRISM consistently outperforms baseline methods such as Mem0 and Memory-R1 across different model sizes, while online experiments further demonstrate its effectiveness in real-world smart home deployment.


\bibliographystyle{ACM-Reference-Format}
\bibliography{reference}

@article{king2024sasha,
  title={Sasha: creative goal-oriented reasoning in smart homes with large language models},
  author={King, Evan and Yu, Haoxiang and Lee, Sangsu and Julien, Christine},
  journal={Proceedings of the ACM on Interactive, Mobile, Wearable and Ubiquitous Technologies},
  volume={8},
  number={1},
  pages={1--38},
  year={2024},
  publisher={ACM New York, NY, USA}
}

@article{rivkin2024aiot,
  title={Aiot smart home via autonomous llm agents},
  author={Rivkin, Dmitriy and Hogan, Francois and Feriani, Amal and Konar, Abhisek and Sigal, Adam and Liu, Xue and Dudek, Gregory},
  journal={IEEE Internet of Things Journal},
  year={2024},
  publisher={IEEE}
}

@article{yin2024harmony,
  title={Harmony: A home agent for responsive management and action optimization with a locally deployed large language model},
  author={Yin, Ziqi and Zhang, Mingxin and Kawahara, Daisuke},
  journal={arXiv preprint arXiv:2410.14252},
  year={2024}
}

@article{li2025homebench,
  title={HomeBench: Evaluating LLMs in Smart Homes with Valid and Invalid Instructions Across Single and Multiple Devices},
  author={Li, Silin and Guo, Yuhang and Yao, Jiashu and Liu, Zeming and Wang, Haifeng},
  journal={arXiv preprint arXiv:2505.19628},
  year={2025}
}

@inproceedings{zhong2024memorybank,
  title={Memorybank: Enhancing large language models with long-term memory},
  author={Zhong, Wanjun and Guo, Lianghong and Gao, Qiqi and Ye, He and Wang, Yanlin},
  booktitle={Proceedings of the AAAI Conference on Artificial Intelligence},
  volume={38},
  number={17},
  pages={19724--19731},
  year={2024}
}

@article{packer2023memgpt,
  title={MemGPT: Towards LLMs as Operating Systems.},
  author={Packer, Charles and Fang, Vivian and Patil, Shishir\_G and Lin, Kevin and Wooders, Sarah and Gonzalez, Joseph\_E},
  year={2023},
  publisher={ArXiv}
}

@article{chhikara2025mem0,
  title={Mem0: Building production-ready ai agents with scalable long-term memory},
  author={Chhikara, Prateek and Khant, Dev and Aryan, Saket and Singh, Taranjeet and Yadav, Deshraj},
  journal={arXiv preprint arXiv:2504.19413},
  year={2025}
}

@article{wang2025mem,
  title={Mem-$\{$$\backslash$alpha$\}$: Learning memory construction via reinforcement learning},
  author={Wang, Yu and Takanobu, Ryuichi and Liang, Zhiqi and Mao, Yuzhen and Hu, Yuanzhe and McAuley, Julian and Wu, Xiaojian},
  journal={arXiv preprint arXiv:2509.25911},
  year={2025}
}

@article{yan2025memory,
  title={Memory-r1: Enhancing large language model agents to manage and utilize memories via reinforcement learning},
  author={Yan, Sikuan and Yang, Xiufeng and Huang, Zuchao and Nie, Ercong and Ding, Zifeng and Li, Zonggen and Ma, Xiaowen and Kersting, Kristian and Pan, Jeff Z and Sch{\"u}tze, Hinrich and others},
  journal={arXiv preprint arXiv:2508.19828},
  year={2025}
}

@inproceedings{maharana-etal-2024-evaluating,
    title = "Evaluating Very Long-Term Conversational Memory of {LLM} Agents",
    author = "Maharana, Adyasha  and
      Lee, Dong-Ho  and
      Tulyakov, Sergey  and
      Bansal, Mohit  and
      Barbieri, Francesco  and
      Fang, Yuwei",
    editor = "Ku, Lun-Wei  and
      Martins, Andre  and
      Srikumar, Vivek",
    booktitle = "Proceedings of the 62nd Annual Meeting of the Association for Computational Linguistics (Volume 1: Long Papers)",
    month = aug,
    year = "2024",
    address = "Bangkok, Thailand",
    publisher = "Association for Computational Linguistics",
    url = "https://aclanthology.org/2024.acl-long.747/",
    doi = "10.18653/v1/2024.acl-long.747",
    pages = "13851--13870"
}

@article{wu2024longmemeval,
  title={Longmemeval: Benchmarking chat assistants on long-term interactive memory},
  author={Wu, Di and Wang, Hongwei and Yu, Wenhao and Zhang, Yuwei and Chang, Kai-Wei and Yu, Dong},
  journal={arXiv preprint arXiv:2410.10813},
  year={2024}
}

@article{huang2025mem,
  title={Mem-PAL: Towards Memory-based Personalized Dialogue Assistants for Long-term User-Agent Interaction},
  author={Huang, Zhaopei and Dai, Qifeng and Wu, Guozheng and Wu, Xiaopeng and Chen, Kehan and Yu, Chuan and Li, Xubin and Ge, Tiezheng and Wang, Wenxuan and Jin, Qin},
  journal={arXiv preprint arXiv:2511.13410},
  year={2025}
}

@inproceedings{tan-etal-2025-membench,
    title = "{M}em{B}ench: Towards More Comprehensive Evaluation on the Memory of {LLM}-based Agents",
    author = "Tan, Haoran  and
      Zhang, Zeyu  and
      Ma, Chen  and
      Chen, Xu  and
      Dai, Quanyu  and
      Dong, Zhenhua",
    editor = "Che, Wanxiang  and
      Nabende, Joyce  and
      Shutova, Ekaterina  and
      Pilehvar, Mohammad Taher",
    booktitle = "Findings of the Association for Computational Linguistics: ACL 2025",
    month = jul,
    year = "2025",
    address = "Vienna, Austria",
    publisher = "Association for Computational Linguistics",
    url = "https://aclanthology.org/2025.findings-acl.989/",
    doi = "10.18653/v1/2025.findings-acl.989",
    pages = "19336--19352",
    ISBN = "979-8-89176-256-5"
}

@article{ai2025memorybench,
  title={MemoryBench: A Benchmark for Memory and Continual Learning in LLM Systems},
  author={Ai, Qingyao and Tang, Yichen and Wang, Changyue and Long, Jianming and Su, Weihang and Liu, Yiqun},
  journal={arXiv preprint arXiv:2510.17281},
  year={2025}
}

@article{shao2025deepseekmath,
  title={Deepseekmath-v2: Towards self-verifiable mathematical reasoning},
  author={Shao, Zhihong and Luo, Yuxiang and Lu, Chengda and Ren, ZZ and Hu, Jiewen and Ye, Tian and Gou, Zhibin and Ma, Shirong and Zhang, Xiaokang},
  journal={arXiv preprint arXiv:2511.22570},
  year={2025}
}

@article{team2025kimi,
  title={Kimi k2: Open agentic intelligence},
  author={Team, Kimi and Bai, Yifan and Bao, Yiping and Chen, Guanduo and Chen, Jiahao and Chen, Ningxin and Chen, Ruijue and Chen, Yanru and Chen, Yuankun and Chen, Yutian and others},
  journal={arXiv preprint arXiv:2507.20534},
  year={2025}
}

@article{yu2025rlpr,
  title={RLPR: Extrapolating RLVR to General Domains without Verifiers},
  author={Yu, Tianyu and Ji, Bo and Wang, Shouli and Yao, Shu and Wang, Zefan and Cui, Ganqu and Yuan, Lifan and Ding, Ning and Yao, Yuan and Liu, Zhiyuan and others},
  journal={arXiv preprint arXiv:2506.18254},
  year={2025}
}

@article{wang2025icpo,
  title={ICPO: Intrinsic Confidence-Driven Group Relative Preference Optimization for Efficient Reinforcement Learning},
  author={Wang, Jinpeng and Li, Chao and Ye, Ting and Zhang, Mengyuan and Liu, Wei and Luan, Jian},
  journal={arXiv preprint arXiv:2511.21005},
  year={2025}
}

@article{tang2025beyond,
  title={Beyond Verifiable Rewards: Scaling Reinforcement Learning for Language Models to Unverifiable Data},
  author={Tang, Yunhao and Wang, Sid and Madaan, Lovish and Munos, R{\'e}mi},
  journal={arXiv preprint arXiv:2503.19618},
  year={2025}
}

@article{su2025crossing,
  title={Crossing the Reward Bridge: Expanding RL with Verifiable Rewards Across Diverse Domains},
  author={Su, Yi and Yu, Dian and Song, Linfeng and Li, Juntao and Mi, Haitao and Tu, Zhaopeng and Zhang, Min and Yu, Dong},
  journal={arXiv preprint arXiv:2503.23829},
  year={2025}
}

@inproceedings{10.5555/3692070.3693141,
author = {Lee, Harrison and Phatale, Samrat and Mansoor, Hassan and Mesnard, Thomas and Ferret, Johan and Lu, Kellie and Bishop, Colton and Hall, Ethan and Carbune, Victor and Rastogi, Abhinav and Prakash, Sushant},
title = {RLAIF vs. RLHF: scaling reinforcement learning from human feedback with AI feedback},
year = {2024},
publisher = {JMLR.org},
booktitle = {Proceedings of the 41st International Conference on Machine Learning},
articleno = {1071},
numpages = {28},
location = {Vienna, Austria},
series = {ICML'24}
}

@inproceedings{peng-etal-2025-verif,
    title = "{V}er{IF}: Verification Engineering for Reinforcement Learning in Instruction Following",
    author = "Peng, Hao  and
      Qi, Yunjia  and
      Wang, Xiaozhi  and
      Xu, Bin  and
      Hou, Lei  and
      Li, Juanzi",
    editor = "Christodoulopoulos, Christos  and
      Chakraborty, Tanmoy  and
      Rose, Carolyn  and
      Peng, Violet",
    booktitle = "Proceedings of the 2025 Conference on Empirical Methods in Natural Language Processing",
    month = nov,
    year = "2025",
    address = "Suzhou, China",
    publisher = "Association for Computational Linguistics",
    url = "https://aclanthology.org/2025.emnlp-main.1542/",
    doi = "10.18653/v1/2025.emnlp-main.1542",
    pages = "30324--30339",
    ISBN = "979-8-89176-332-6"
}

@article{shao2024deepseekmath,
  title={Deepseekmath: Pushing the limits of mathematical reasoning in open language models},
  author={Shao, Zhihong and Wang, Peiyi and Zhu, Qihao and Xu, Runxin and Song, Junxiao and Bi, Xiao and Zhang, Haowei and Zhang, Mingchuan and Li, YK and Wu, Yang and others},
  journal={arXiv preprint arXiv:2402.03300},
  year={2024}
}

@article{yang2025qwen3,
  title={Qwen3 technical report},
  author={Yang, An and Li, Anfeng and Yang, Baosong and Zhang, Beichen and Hui, Binyuan and Zheng, Bo and Yu, Bowen and Gao, Chang and Huang, Chengen and Lv, Chenxu and others},
  journal={arXiv preprint arXiv:2505.09388},
  year={2025}
}

@article{liu2024deepseek,
  title={Deepseek-v3 technical report},
  author={Liu, Aixin and Feng, Bei and Xue, Bing and Wang, Bingxuan and Wu, Bochao and Lu, Chengda and Zhao, Chenggang and Deng, Chengqi and Zhang, Chenyu and Ruan, Chong and others},
  journal={arXiv preprint arXiv:2412.19437},
  year={2024}
}

@misc{qwen2.5,
    title = {Qwen2.5: A Party of Foundation Models},
    url = {https://qwenlm.github.io/blog/qwen2.5/},
    author = {Qwen Team},
    month = {September},
    year = {2024}
}

\clearpage
\onecolumn
\appendix

\section{Ablation on Reward Design.}\label{ablation}
The detailed results of ablation studies under different prefix reward weights and judge granularities are reported in Table \ref{tab:ablation_reward}.
\begin{table*}[h]
  \centering
  \small
  \setlength{\tabcolsep}{6pt}
  \renewcommand{\arraystretch}{1.25}
  \begin{tabular}{l|ccc|ccc|ccc|ccc}
    \hline
    \multirow{2}{*}{\textbf{Setting}}
    & \multicolumn{3}{c|}{\textbf{No-Memory}}
    & \multicolumn{3}{c|}{\textbf{Memory Use}}
    & \multicolumn{3}{c|}{\textbf{Memory State Change}}
    & \multicolumn{3}{c}{\textbf{Overall}} \\
    & F1$\uparrow$ & B1$\uparrow$ & Acc$\uparrow$
    & F1$\uparrow$ & B1$\uparrow$ & Acc$\uparrow$
    & F1$\uparrow$ & B1$\uparrow$ & Acc$\uparrow$
    & F1$\uparrow$ & B1$\uparrow$ & Acc$\uparrow$ \\
    \hline

    prefix $\lambda{=}0.3$
    & \textbf{56.88} & \textbf{54.25} & 52.83
    & \textbf{65.10} & \textbf{62.85} & \textbf{58.64}
    & 81.30 & 76.94 & 72.41
    & \textbf{68.81} & \textbf{65.88} & \textbf{61.95} \\

    \hline

    prefix $\lambda{=}0$
    & 50.76 & 48.89 & 54.72
    & 62.70 & 60.23 & 55.45
    & 80.06 & 76.47 & 74.14
    & 66.25 & 63.53 & 60.93 \\

    prefix $\lambda{=}0.5$
    & 44.42 & 43.23 & 54.72
    & 64.96 & 61.93 & 55.91
    & 78.86 & 75.08 & 73.28
    & 66.31 & 63.30 & 60.93 \\

    prefix $\lambda{=}0.7$
    & 44.96 & 43.43 & 50.94
    & 62.05 & 59.76 & 52.73
    & 80.17 & 76.42 & 70.69
    & 65.13 & 62.50 & 57.84 \\

    \hline

    Additive Multi-Dim Reward (No Veto)
    & 42.46 & 40.15 & 50.94
    & 61.01 & 57.63 & 50.00
    & \textbf{81.69} & \textbf{77.38} & \textbf{74.14}
    & 64.65 & 61.14 & 57.33 \\

    \hline
  \end{tabular}
  \caption{Ablation study on reward design (4B models).
    We compare different prefix reward weights $\lambda$ and judge granularities.
    The first row shows the full model configuration ($\lambda{=}0.3$ with multi-dimensional judges), while the remaining rows correspond to ablations.
  All metrics are higher-is-better. Best results among ablations are highlighted in bold.}
  \label{tab:ablation_reward}
\end{table*}

\section{Similarity}\label{similarity}
Let the model-generated command be represented as a token sequence
$Y = (y_1, \dots, y_n)$, and the ground-truth as $Y^* = (y^*_1, \dots, y^*_m)$. We first compute token-level precision and recall:
\begin{align}
  \text{Precision} &= \frac{|Y \cap Y^*|}{|Y|}, \\
  \text{Recall} &= \frac{|Y \cap Y^*|}{|Y^*|}.
\end{align}
The token-level F1 score is then computed as:
\begin{align}
  \text{F1}
  &= \frac{2 \cdot \text{Precision} \cdot \text{Recall}}
  {\text{Precision} + \text{Recall}}.
\end{align}
BLEU-1 (B1) measures generation quality from the unigram matching perspective. Its core component is the modified unigram precision:
\begin{align}
  p_1
  &= \frac{\sum_{w \in Y} \min\!\left(\text{count}_Y(w), \text{count}_{Y^*}(w)\right)}
  {\sum_{w \in Y} \text{count}_Y(w)}.
\end{align}
The final BLEU-1 score is obtained by applying a brevity penalty (BP):
\begin{align}
  \text{BLEU-1}
  &= \text{BP} \cdot p_1, \\
  \text{BP}
  &=
  \begin{cases}
    1, & |Y| > |Y^*|, \\
    \exp\!\left(1 - \dfrac{|Y^*|}{|Y|}\right), & |Y| \le |Y^*|.
  \end{cases}
\end{align}

\section{Prompts}\label{prompts}
\subsection{Device Control Prompt}

\begin{CJK}{UTF8}{gbsn}

  \begin{tcolorbox}[
      title={Device Control Prompt},
      colback=gray!5,
      colframe=gray!60,
      fonttitle=\bfseries,
      breakable,
      left=6pt,
      right=6pt,
      top=6pt,
      bottom=6pt
    ]

    你是一个智能家居记忆 Agent。给定以下输入：
    - 用户当前指令（query）
    - 最近 1–2 轮对话历史（history）
    - 当前长期记忆（memory）
    - 当前入口房间（enter\_room）

    你的任务是根据输入，选择且仅选择以下三种输出类型之一：
    【记忆】、【改写】、【不改写】。

    \textbf{记忆}
    当且仅当用户在 query 或 history 中显式要求系统“记住”某条规则时触发。
    仅允许记忆与智能家居设备控制、查询或模式相关的内容。

    \textbf{改写}
    当当前 query 可以被已有记忆补充或替换，以更准确反映用户长期偏好时执行改写。
    改写需满足房间范围一致性。

    \textbf{不改写}
    当不存在可用记忆，或记忆与当前 query 不匹配时，选择不改写。

    三类输出具有严格优先级：
    若判定为【记忆】，则不再考虑【改写】或【不改写】。

    请仅输出最终结果内容，不要输出解释。

    \textbf{Example:}

    memory: []
    history: []
    query: 帮我记住我以后每次打开空调的时候，都要设置25度
    enter\_room: 客厅

    output: 记忆：好的，已帮您记住"打开空调，就要设置25度"

  \end{tcolorbox}

\end{CJK}

\subsection{PRISM Reward Judge Prompts}

We employ multiple independent LLM-based judges to evaluate output correctness along different consistency dimensions.
Each judge produces a binary decision (\texttt{Y} or \texttt{N}) indicating whether the model output is consistent with the ground truth on the corresponding dimension.

\begin{CJK}{UTF8}{gbsn}
  \begin{tcolorbox}[breakable,
      colback=gray!5,
      colframe=black!60,
      title=Judge~1: Core Semantic Consistency
    ]
\begin{verbatim}
你是一名智能家居记忆任务的评估专家。
你的任务是判断模型输出在核心语义上是否与标准答案一致。

请重点关注：
- 设备主体是否一致（如空调、洗衣机、灯具等）
- 操作或功能是否一致（开关、模式、参数设置等）
- 参数是否存在缺失、错误或多余（温度、风速、水温等）
- 作用范围是否一致（具体房间、全部设备、默认房间）

若模型输出在上述关键信息上与标准答案一致，输出 Y；
否则输出 N。

请只输出 Y 或 N，不要输出解释。
\end{verbatim}
  \end{tcolorbox}
\end{CJK}

\begin{CJK}{UTF8}{gbsn}
  \begin{tcolorbox}[breakable,
      colback=gray!5,
      colframe=black!60,
      title=Judge~2: Rule and Intent Consistency
    ]
\begin{verbatim}
你是一名智能家居记忆任务的评估专家。
你的任务是判断模型输出在规则结构和用户意图层面
是否与标准答案保持一致。

请重点关注：
- 用户的规则意图是否被正确保留
- 模型是否错误地将组合动作改写为条件触发逻辑，或相反
- 输出在语义关系或执行逻辑上是否发生实质性偏移

若模型输出在规则逻辑与用户意图上与标准答案一致，
输出 Y；否则输出 N。

请只输出 Y 或 N，不要输出解释。
\end{verbatim}
  \end{tcolorbox}
\end{CJK}

\begin{CJK}{UTF8}{gbsn}
  \begin{tcolorbox}[breakable,
      colback=gray!5,
      colframe=black!60,
      title=Judge~3: Memory Rejection Consistency
    ]
\begin{verbatim}
你是一名智能家居记忆任务的评估专家。
你的任务是判断模型输出在“不记忆”与拒绝规则上
是否与标准答案一致。

请重点关注：
- 模型是否仍错误写入记忆
- 是否违反“非家电设备内容不记忆”的规则
- 是否违反“仅具有临时意义的请求不记忆”的规则
- 模型是否在应当不改写时仍触发了记忆或改写

若模型输出在不记忆与拒绝规则上与标准答案一致，
输出 Y；否则输出 N。

请只输出 Y 或 N，不要输出解释。
\end{verbatim}
  \end{tcolorbox}
\end{CJK}

\subsection{Unified Consistency Judge Prompt}
\begin{CJK}{UTF8}{gbsn}
  \begin{tcolorbox}[breakable,
      colback=gray!5,
      colframe=black!60,
      title=Unified LLM-Based Consistency Judge
    ]
\begin{verbatim}
你是一名智能家居语音助手中用户记忆任务的评估专家。
你的任务是判断模型预测结果是否与标准答案在语义与行为上保持一致。

给定输入：
- 用户请求（REQUEST）
- 标准答案（GROUND_TRUTH）
- 模型预测结果（PREDICT_OUTPUT）

请重点关注：
- 设备主体是否一致（如空调、洗衣机、灯具等）
- 操作或功能是否一致（开关、模式、参数设置等）
- 参数是否存在缺失、错误或多余（温度、风速、水温等）
- 作用范围是否一致（具体房间、全部设备、默认房间）
- 用户的规则意图是否被正确保留
- 模型是否错误地将组合动作改写为条件触发逻辑，或相反
- 输出在语义关系或执行逻辑上是否发生实质性偏移
- 模型是否仍错误写入记忆
- 是否违反“非家电设备内容不记忆”的规则
- 是否违反“仅具有临时意义的请求不记忆”的规则
- 模型是否在应当不改写时仍触发了记忆或改写

如果模型预测结果在上述方面均与标准答案一致，输出 Y；否则输出 N。

请只输出 Y 或 N，不要输出解释。
\end{verbatim}
  \end{tcolorbox}
\end{CJK}

\subsection{LLM-Based Evaluation Judge}
\begin{CJK}{UTF8}{gbsn}
  \begin{tcolorbox}[breakable,
      colback=gray!5,
      colframe=black!70,
      title=Evaluation Prompt for Memory-Related Tasks
    ]
\begin{verbatim}
# 角色定义
你是一个智能家电语音助手领域用户记忆功能的评估专家，
你将公正地评估关于用户显示记忆的记录(memory add)
和记忆使用(memory use)任务的准确率。

# 任务描述
你需要先理解下面记忆任务说明和判断标准，
然后结合请求输入 (REQUEST)、标准答案（GROUND_TRUTH）
来判断模型预测结果 (PREDICT_OUTPUT) 是否正确。

# Output要求
整体输出格式如下:
<output>true/false</output>
<explain>{错误时的解释}</explain>

其中：
- output 表示模型预测结果是否正确（true 为正确，false 为错误）
- explain 仅在 false 时给出错误原因

# 记忆任务说明

## 记忆功能介绍
记忆功能是家电智能助手中记录用户个性化(memory add)
和使用时(memory use)实现用户个性化的关键能力。
核心包括用户记忆记录和带记忆推理两方面。

## 记忆记录 (memory add)
记忆记录功能用于识别用户请求(query & history)
中与设备相关的显性偏好表达。

- “显性”表达：用户明确提到“记住”“以后”“留心下”等；
  隐式偏好暂不记忆。
- “记忆”范围：仅限设备控制、家电问答、设备查询等内容；
  非设备类（如个人属性、临时信息）不记忆。

记忆记录输出说明：
- 若应记忆，输出：
  好的，已帮你记住{记忆内容的复述}
- 若为设备定时控制类，严格输出：
  不改写

## 记忆使用 (memory use)
在常规请求中，判断 Query 是否与 Memory 相关：
- 若相关，基于 Memory 补充 Query，输出：
  改写：{改写后的新 query}
- 若不相关，不使用 Memory，输出：
  不改写

# 判断标准
1. 若 PREDICT_OUTPUT 与 GROUND_TRUTH 的类型不一致，
   直接判定为错误（false）。
2. 类型一致时，检查是否存在关键信息缺失或不一致：
   - 设备主体、功能、参数是否缺失或多余
   - 范围是否一致（房间、设备集合等）
3. 记忆回复话术格式不作为强约束，
   只要关键信息一致即可。
\end{verbatim}
  \end{tcolorbox}
\end{CJK}

\section{Training Hyperparameters}\label{parameters}

We adopt a two-stage training pipeline consisting of supervised fine-tuning (SFT) followed by reinforcement learning (RL).
The SFT stage initializes the model with basic memory-related behaviors, while the RL stage further aligns the model using the proposed PRISM reward.
The training hyperparameters for both stages are summarized in Table~\ref{tab:training_hparams}.

\begin{table}[t]
  \centering
  \small
  \setlength{\tabcolsep}{6pt}
  \renewcommand{\arraystretch}{1.15}
  \begin{tabular}{l c c}
    \toprule
    \textbf{Hyperparameter} & \textbf{SFT} & \textbf{GRPO with PRISM} \\
    \midrule
    Base model & Qwen3 (4B / 8B) & Qwen3 (4B / 8B) \\
    Training epochs & 3 & 4 \\
    Train batch size & 16 & 8 \\
    Learning rate & $1 \times 10^{-5}$ & $5 \times 10^{-7}$ \\
    Optimizer & AdamW & -- \\
    PPO mini-batch size & -- & 4 \\
    PPO micro-batch (per GPU) & -- & 2 \\
    KL penalty coefficient & -- & 0.03 \\
    KL type & -- & Low-variance KL \\
    Rollout samples ($n$) & -- & 8 \\
    Sampling temperature & -- & 0.8 \\
    Top-$p$ & -- & 0.9 \\
    Max prompt length & 8192 & 8192 \\
    Max response length & 512 & 512 \\
    Gradient checkpointing & Enabled & Enabled \\
    \bottomrule
  \end{tabular}
  \caption{Training hyperparameters for supervised fine-tuning (SFT) and reinforcement learning (RL) stages.}
  \label{tab:training_hparams}
\end{table}

\section{Error Analysis}\label{ea}
We conduct a systematic error analysis based on the evaluation results of \textbf{Gemini-2.5-Pro} on the MemHome test set. Several error cases are presented.

\textbf{Incomplete Memory Construction in Multi-turn Dialogues}. In some cases, the model is required to aggregate information from multiple dialogue turns to construct a complete memory entry. We observe that critical information may be omitted during this process.

\begin{CJK}{UTF8}{gbsn}
  \begin{tcolorbox}[
      colback=gray!5,
      colframe=gray!60,
      breakable,
      left=6pt,right=6pt,top=6pt,bottom=6pt
    ]
    \textbf{Dialogue history:}
    用户：现在几点了
    助手：现在是下午3点15分。
    用户：打开客厅空调，有点闷
    助手：好的，已为您打开客厅空调。
    用户：请你记一下那个
    助手：请问你要我为你记住什么呢？

    \medskip
    \textbf{Current query:}
    以后我说“太闷了”就给我开送风模式

    \medskip
    \textbf{Memory state:} 打开客厅照明时只开射灯

    \medskip
    \textbf{Ground truth:}
    记忆：好的，已帮您记住说“太闷了”时，就为你打开客厅空调的送风模式。

    \medskip
    \textbf{Model output:}
    记忆：好的，已帮您记住用户说“太闷了”就开送风模式。
  \end{tcolorbox}
\end{CJK}

\textbf{Failure under Colloquial or Implicit Expressions}. Highly colloquial or implicit expressions may prevent the model from correctly associating the query with existing memory, or lead to rewritten outputs with missing key information.

\begin{CJK}{UTF8}{gbsn}
  \begin{tcolorbox}[
      colback=gray!5,
      colframe=gray!60,
      breakable
    ]
    \textbf{Dialogue history:}
    用户：灯光
    助手：需要打开还是调节亮度还是要关了？

    \medskip
    \textbf{Current query:}
    是赶紧调下

    \medskip
    \textbf{Memory state:}
    调节客厅灯光亮度时默认为60\%暖白

    \medskip
    \textbf{Ground truth:}
    改写：将客厅灯光亮度调节为60\%暖白。

    \medskip
    \textbf{Model output:}
    不改写
  \end{tcolorbox}
\end{CJK}

\begin{CJK}{UTF8}{gbsn}
  \begin{tcolorbox}[
      colback=gray!5,
      colframe=gray!60,
      breakable
    ]
    \textbf{Current query:}
    风扇吹自然风

    \medskip
    \textbf{Memory state:}
    说自然风风扇就切换到自然风模式并摇头

    \medskip
    \textbf{Ground truth:}
    改写：风扇切换到自然风模式并摇头。

    \medskip
    \textbf{Model output:}
    改写：切换到自然风模式并摇头。
  \end{tcolorbox}
\end{CJK}

\textbf{Incorrect Memory Usage} The model may incorrectly apply existing memory when the current query does not require memory involvement.

\begin{CJK}{UTF8}{gbsn}
  \begin{tcolorbox}[
      colback=gray!5,
      colframe=gray!60,
      breakable
    ]
    \textbf{Dialogue history:}
    用户：空调模式
    助手：请问你调节空调什么模式呢？

    \medskip
    \textbf{Current query:}
    太热了，降一点

    \medskip
    \textbf{Memory state:}
    客厅空调的平常设置为除湿模式，温度二十六度，风速中

    \medskip
    \textbf{Ground truth:}
    不改写

    \medskip
    \textbf{Model output:}
    改写：设置客厅空调为除湿模式，温度二十六度，风速中。
  \end{tcolorbox}
\end{CJK}

Overall, errors mainly stem from multi-memory disambiguation, weak or implicit memory triggers, and unstable boundary judgments for memory updates. These failure modes reflect structural challenges in memory-driven decision making rather than surface-level language generation issues.

\end{document}